%% file: main.tex
\documentclass[10pt,twocolumn,letterpaper]{article}
\input{defs}

\usepackage{iccv}
\usepackage{times}
\usepackage{epsfig}
\usepackage{graphicx}
\usepackage{amsmath}
\usepackage{amssymb}
\usepackage{booktabs}       
\usepackage{amsfonts}       
\usepackage{nicefrac}       
\usepackage{microtype}      
\usepackage{graphicx}
\usepackage{multirow}
\usepackage{authblk}
\usepackage[font={small}]{caption}

\newenvironment{packed_item}{
\begin{itemize}
  \setlength{\itemsep}{1pt}
  \setlength{\parskip}{0pt}
  \setlength{\parsep}{0pt}
}{\end{itemize}}

\newenvironment{packed_enum}{
\begin{enumerate}
  \setlength{\itemsep}{1pt}
  \setlength{\parskip}{0pt}
  \setlength{\parsep}{0pt}
}{\end{enumerate}}

\usepackage[pagebackref=true,breaklinks=true,letterpaper=true,colorlinks,bookmarks=false]{hyperref}

\def\iccvPaperID{1523} 

\iccvfinalcopy
\ificcvfinal\fi
\begin{document}

\title{Low-shot Visual Recognition by Shrinking and Hallucinating Features}
\author{Bharath Hariharan}
\author{Ross Girshick}
\affil{Facebook AI Research (FAIR)}

\twocolumn[{%
\renewcommand\twocolumn[1][]{#1}%
\maketitle
\vspace{-1cm}
\begin{center}
    \centering 
    \includegraphics[width=0.8\textwidth]{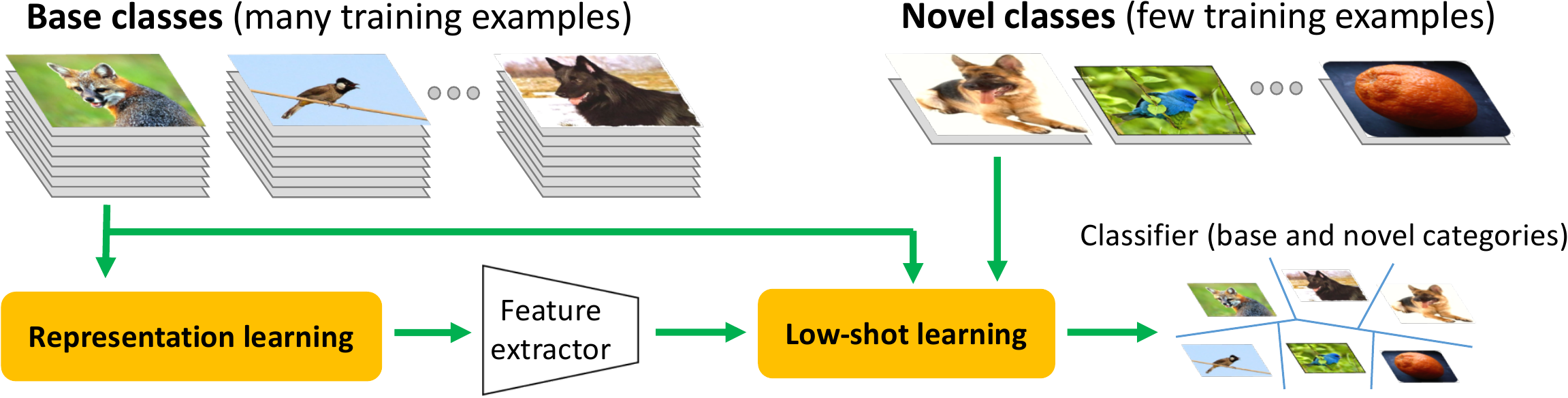}
    \captionof{figure}{\textbf{Our low-shot learning benchmark in two phases: representation learning and low-shot learning.} Modern recognition models use large labeled datasets like ImageNet to build good visual representations and train strong classifiers (\emph{representation learning}).
    However, these datasets only contain a fixed set of classes.
    In many realistic scenarios, once deployed, the model might encounter novel classes that it also needs to recognize, but with very few training examples available (\emph{low-shot learning}).
    We present two ways of significantly improving performance in this scenario: (1) a novel loss function for representation learning that leads to better visual representations that generalize well, and (2) a method for hallucinating additional examples for the data-starved novel classes.}
    \label{teaser}
\end{center}%
}]

\begin{abstract}
Low-shot visual learning---the ability to recognize novel object categories from very few examples---is a hallmark of human visual intelligence.
Existing machine learning approaches fail to generalize in the same way.
  To make progress on this foundational problem, we present a low-shot learning benchmark on complex images that mimics challenges faced by recognition systems in the wild. 
  We then propose (1) representation regularization techniques, and (2) techniques to hallucinate additional training examples for data-starved classes. 
  Together, our methods improve the effectiveness of convolutional networks in low-shot learning, improving the one-shot accuracy on novel classes by 2.3$\times$ on the challenging ImageNet dataset.
\end{abstract}
\input{intro.tex}
\input{relwork.tex}

\input{evaluation_protocol.tex}
\input{approach_gen.tex}
\input{approach.tex}
\input{experiments.tex}

\input{conclusion.tex}
{\small
\bibliographystyle{ieee}
\bibliography{main.bib}
}

\end{document}

%% file: defs.tex
\usepackage{color}
\newcommand{\firstrank}[1]{\textbf{\textcolor{blue}{#1}}}
\newcommand{\secondrank}[1]{\textit{\textcolor{red}{#1}}}

\definecolor{rbg_color}{rgb}{.6,.25,.5}
\definecolor{bh_color}{rgb}{1,.5,0}
\definecolor{todo_color}{rgb}{.8,.8,.1}
\definecolor{new_color}{rgb}{.9,.1,.1}

\newcommand{\base}[0]{\mathit{base}}
\newcommand{\novel}[0]{\mathit{novel}}

\newcommand{\bphi}{\mbox{\boldmath $\phi$}}
\newcommand{\feat}{\bphi}

\usepackage{amsopn}

%% file: intro.tex
\section{Introduction}
\label{intro}

Recently, error rates on benchmarks like ImageNet \cite{imagenet_cvpr09} have been halved, and then halved again.
These gains come from deep convolutional networks (ConvNets) that learn rich feature representations \cite{krizhevsky2012imagenet}.
It is now clear that if an application has an a priori fixed set of visual concepts and thousands of examples per concept, an effective way to build an object recognition system is to train a deep ConvNet.
But what if these assumptions are not satisfied and the network must learn novel categories from very few examples?

The ability to perform \emph{low-shot learning}---learning novel concepts from very few examples---is a hallmark of the human visual system.
We are able to do this not only for natural object categories such as different kinds of animals, but also for synthetic objects that are unlike anything we've seen before~\cite{Schmidt2009}.
In contrast, in spite of significant improvements in recognition performance, computational recognition approaches fail to generalize well from few examples~\cite{LakeScience2015}.
Our goal in this paper is to make progress towards imparting this human ability to modern recognition systems.

Our first contribution is a low-shot learning benchmark based on the challenging ImageNet1k dataset.
As shown in Figure 1, our benchmark is implemented in two phases.
In the \emph{representation learning phase}, the learner tunes its feature representation on a set of base classes that have many training instances.
In the \emph{low-shot learning phase}, the learner is exposed to a set of novel classes with only a few examples per class and must learn a classifier over the \emph{joint} label space of base and novel classes.
This benchmark simulates a scenario in which the learner is deployed in the wild and has to quickly learn novel concepts it encounters from very little training data.
Unlike previous low-shot learning tests (\eg, \cite{FeiFeiTPAMI2006,LakeScience2015}) we measure the learner's accuracy on both the base and novel classes.
This provides a sanity check that accuracy gains on novel classes do not come at the expense of a large loss in performance on the base classes.
This evaluation protocol follows the standard way that image classifiers are evaluated on popular benchmarks like ImageNet, thus easing the comparison of progress on low-shot learning to the typical data-rich scenario.

Next, we investigate how to improve the learner's performance on the benchmark.
We build on the intuition that certain modes of intra-class variation generalize across categories (\eg, pose transformations).
We present a way of ``hallucinating'' additional examples for novel classes by transferring modes of variation from the base classes.
These additional examples improve the one-shot top-5 accuracy on novel classes by \emph{15 points} (absolute) while also maintaining accuracy on the base classes.

Finally, we show that the feature representation learnt in the first phase has a large impact on low-shot generalization ability.
Specifically, we formulate a loss function that penalizes the difference between classifiers learnt on large and small datasets, and then draw connections between this loss and regularization of \emph{feature activations}.
We show that simply regularizing feature activations can increase one-shot, top-5 accuracy on novel classes by \emph{9 points} (absolute) without harming base class performance. 
Combining this better representation with the hallucination strategy pushes our improvement up to \emph{18 points} above the baseline.

%
%

%% file: relwork.tex
\section{Related work}
\textbf{One-shot and low-shot learning.}
One class of approaches to one-shot learning uses generative models of appearance that tap into a global~\cite{FeiFeiTPAMI2006} or a supercategory-level~\cite{Salakhutdinov2012} prior.
Generative models based on strokes~\cite{LakeNIPS2013} or parts~\cite{WongICCV2015} have shown promise in restricted domains such as hand-written characters~\cite{Lecun1998, LakeScience2015}.
They also work well in datasets without much intra-class variation or clutter, such as Caltech 101~\cite{FeiFeiTPAMI2006}.
Dixit \etal~\cite{DixitCVPR2017} leverage a corpus with attribute annotations to generate additional examples by varying attributes.
We also propose a way to generate additional examples, but our model does not use supervision.
A similar approach to synthesizing additional examples by transforming existing ones is presented in early work by Miller \etal~\cite{MillerCVPR2000}.
Our approach generalizes this to realistic, generic image categories and is non-parametric. 

Jia \etal~\cite{JiaICCV2013} present a promising alternative to generation using Bayesian reasoning to infer an object category from a few examples; however, in~\cite{JiaICCV2013} the full, large-scale training set is available during training.

Among discriminative approaches, early work attempted to use a single image of the novel class to adapt classifiers from similar base classes~\cite{BartCVPR2005, OpeltCVPR2006} using simple hand-crafted features.
Bertinetto \etal~\cite{BertinettoNIPS2016} regress from single examples to a classifiers, while Wang and Hebert~\cite{WangECCV2016} regress from classifiers trained on small datasets to classifiers trained on large datasets.
Recent ``meta-learning" techniques learn to directly map training sets and test examples to classification outputs~\cite{VinyalsNIPS2016, FinnICML2017, RaviICLR2017}.
We compare favorably with these approaches in our experiments.

Amongst representation learning approaches, metric learning, such as the triplet loss~\cite{TaigmanCVPR2015, SchroffCVPR2015, FinkNIPS2005} or siamese networks~\cite{KochICMLW2015,HadsellCVPR2006}, has been used to automatically learn feature representations where objects of the same class are closer together.
Such approaches have shown benefits in face identification~\cite{TaigmanCVPR2015}.
On benchmarks involving more general Internet imagery, such as ImageNet~\cite{imagenet_cvpr09}, these methods perform worse than simple classification baselines~\cite{RippelICLR2016}, and it is unclear if they can benefit low-shot learning.

\textbf{Zero-shot learning.}
Zero-shot recognition uses textual or attribute-level descriptions of object classes to train classifiers.
While this problem is different than ours, the motivation is the same: to reduce the amount of data required to learn classifiers.
One line of work uses hand-designed attribute descriptions that are provided to the system for the novel categories~\cite{RomeraICML2015, LampertTPAMI2014, FarhadiCVPR2010}.
Another class of approaches embeds images into word embedding spaces learnt using large text corpora, so that classifiers for novel concepts can be obtained simply from the word embedding of the concept~\cite{FromeNIPS2013, SocherNIPS2013, NorouziICLR2014, ZhangICCV2015}.
A final class of approaches attempts to directly regress to image classifiers from textual descriptions ~\cite{ElhoseinyICCV2013, BaICCV2015} or from prototypical images of the category~\cite{JetleyBMVC2015}.
Similar to our benchmark, Chao \etal~\cite{ChaoECCV2016} propose that zero-shot learning evaluation should also include the training categories that do have examples. 
We believe this evaluation style is good for both zero and low-shot learning.


\textbf{Transfer learning.}
The ability to learn novel classes quickly is one of the main motivations for multitask and transfer learning.
Thrun's classic paper convincingly argues that ``learning the $n$-th task should be easier than learning the first,'' with ease referring to sample complexity~\cite{ThrunNIPS1996}.
However, recent transfer learning research has mostly focussed on the scenario where large amounts of training data are available for novel classes.
For that situation, the efficacy of pre-trained ConvNets for extracting features is well known~\cite{DonahueICML2014, OquabCVPR2014, RazavianCVPRW2014}.
There is also some analysis on what aspects of ImageNet training aid this transfer~\cite{AgrawalECCV2014, AzizpourCVPRW2015}.
For faces, Taigman \etal~\cite{TaigmanCVPR2015} find that low-dimensional feature representations transfer better on faces and Galanti \etal~\cite{Galanti2016} provide some theoretical justification for this finding.
This work hints at a link between the complexity of the feature representation and its generalizability, a link which we also observe in this paper.
We find that stronger base classifiers generalize better than weaker classifiers (\eg comparing ResNet-10 to ResNet-50 \cite{HeCVPR2016}).
There have also been novel losses proposed explicitly to aid transfer, such as the multiverse loss of Littwin and Wolf~\cite{LittwinArxiv2015}.
Our paper also proposes novel losses designed specifically for low-shot learning.

%% file: evaluation_protocol.tex
\section{A low-shot learning benchmark}
Our goal is to build a benchmark for low-shot learning that mimics situations that arise in practice.
Current recognition systems require days or even weeks of training on expensive hardware to develop good feature representations.
The trained recognition systems may then be deployed as a service to be used by downstream applications.
These downstream applications may need the ability to recognize novel categories, but they may have neither the training data required, nor the infrastructure needed to retrain the models.
Thus, there are two natural phases: in the first phase, we have the data and resources to train sophisticated feature extractors on large labelled datasets, and in the second phase, we want to add additional categories to our repertoire at minimal computational and data cost.

Our low-shot learning benchmark implements a similar setup.
It employs a learner, two training phases, and one testing phase.
The learner is assumed to be composed of a feature extractor and a multi-class classifier.
The benchmark is agnostic to the specific form of each component.

During \emph{representation learning} (training phase one), the learner receives a fixed set of base categories $C_{\base}$, and a dataset $D$ containing a large number of examples for each category in $C_{\base}$. 
The learner uses $D$ to set the parameters of its feature extractor.

In the second phase, which we call \emph{low-shot learning}, the learner is given a set of categories $C^{l}$ that it must learn to distinguish.
$C^{l} = C_{\base} \cup C_{\novel}$ is a mix of base categories $C_{base}$, and unseen novel categories $C_{novel}$. 
For each novel category, the learner has access to only $n$ positive examples, where $n \in \{1,2,5,10,20\}$.
For the base categories, the learner still has access to $D$.
The learner may then use these examples and its feature extractor to set the parameters of its multi-class classifier while also optionally modifying the feature extractor.

In the testing phase, the learnt model predicts labels from the combined label space $C_{\base} \cup C_{\novel}$ on a set of previously unseen test images.
To measure the variability in low-shot learning accuracy, we repeat the low-shot learning and testing phases for 5 trials, each time with a random draw of examples for the novel classes.
We report the mean accuracy and the standard deviation over these trials.





The simplest, and commonly used,  baseline approach is to train a ConvNet with label cross-entropy loss in the representation learning phase and then train a new linear classifier head in the low-shot learning phase.
We now show significant improvements on this baseline, first by a novel strategy of hallucinating additional training examples (Section \ref{sec:approach_gen}) and then by improving the representation itself (Section \ref{sec:approach}).

%% file: approach_gen.tex
\section{Better low-shot learning through generation}
\label{sec:approach_gen}
In the low-shot learning phase, our goal is to train good classifiers for novel categories from only a few examples.
Intuitively, the challenge is that these examples capture very little of the category's intra-class variation.
For instance, if the category is a particular bird species, then we may only have examples of the bird perched on a branch, and none of it in flight.
The classifier might then erroneously conclude that this novel category only consists of perched birds.

However, this mode of variation is common to many bird species, including those we have encountered in the base classes.
From the many base class examples we have seen, we can understand the transformation that relates perched bird images to the image of the corresponding bird in flight, and then use this transformation to ``hallucinate" additional examples for our novel bird category.
If we were given the set of all such category-independent transformations, then we can hallucinate as many new examples for each novel category example as there are transformations.

However, we do not have a pre-defined set of transformations that we can apply.
But we can take a non-parametric approach.
Any two examples $z_1$ and $z_2$ belonging to the same category represent a plausible transformation.
Then, given a novel category example $x$, we want to apply to $x$ the transformation that sent $z_1$ to $z_2$.
That is, we want to complete the transformation ``analogy'' $z_1:z_2::x:\; ?$.

We do this by training a function $G$ that takes as input the concatenated feature vectors of the three examples $[\feat(x), \feat(z_1), \feat(z_2)] $.
It produces as output a ``hallucinated" feature vector (of the same dimensionality as $\feat$), which corresponds to applying the $z_1\rightarrow z_2$ transformation to $x$.
We use an MLP with three fully connected layers for $G$.

We first describe how we train $G$, and then show how we use the generated examples in the low-shot learning phase.

\subsection{Learning to generate new examples}  

\begin{figure}
\centering
\includegraphics[width=\linewidth]{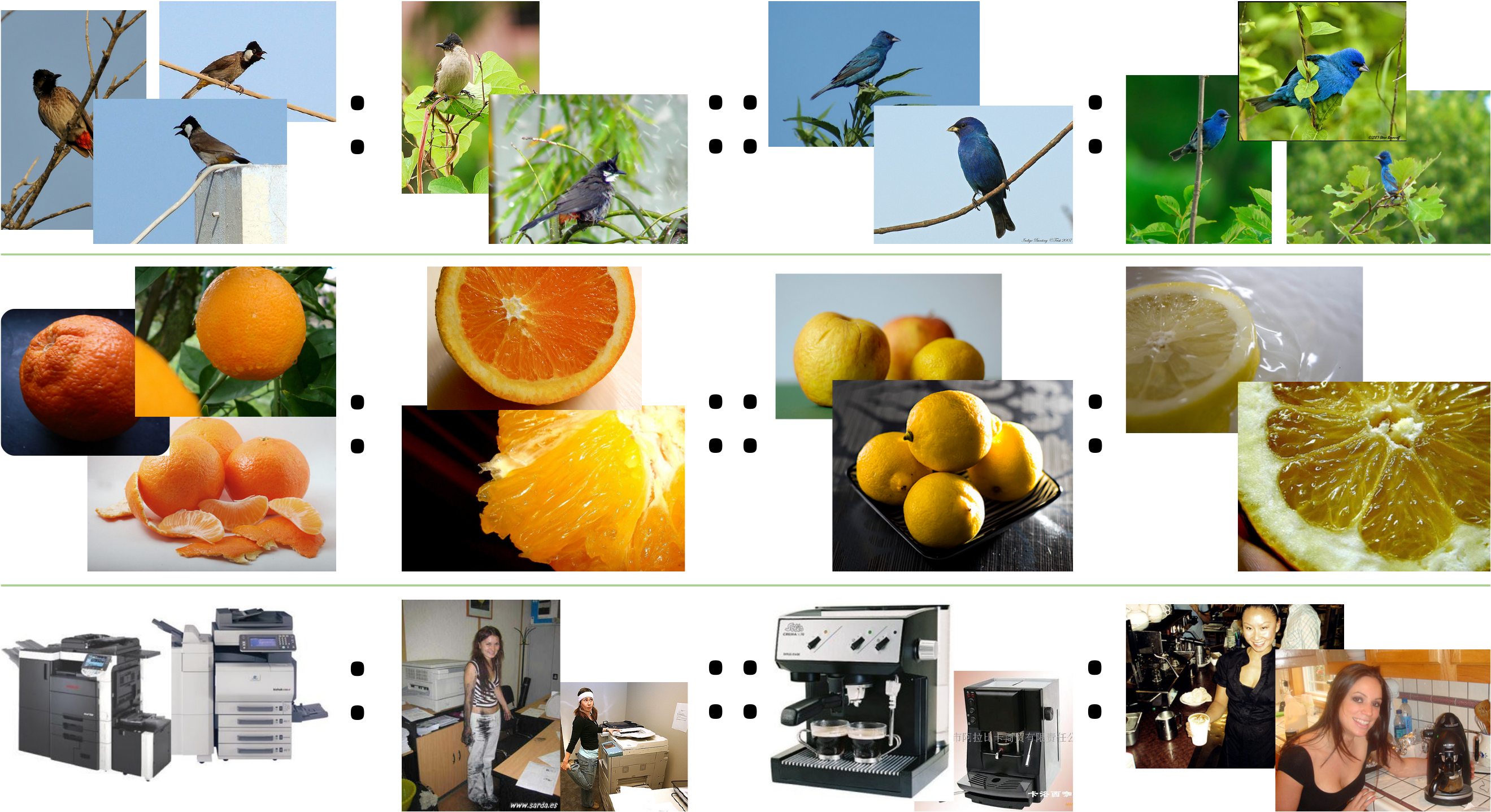}
\caption{Example mined analogies. Each row shows the four image clusters that form the four elements in the analogy. \textbf{Row 1:} birds with a sky backdrop vs birds with greenery in the background. \textbf{Row 2:} whole fruits vs cut fruit. \textbf{Row 3:} machines (printer, coffee making) in isolation vs the same machine operated by a human.}
\label{fig:minedanalogies}
\end{figure}

To train $G$, we first collect a dataset of completed analogies from our base classes.
To do this we first cluster the feature vectors of the examples in each base category into a fixed number of clusters (100). This is to keep computational complexity manageable.
Next, for each pair of centroids $c^a_1, c^a_2$ in one category $a$, we search for another pair of centroids $c^b_1, c^b_2$ from another category $b$, such that the cosine distance between $c^a_1 - c^a_2$ and $c^b_1 - c^b_2$ is minimized. 
We collect all such quadruplets $(c^a_1, c^a_2, c^b_1, c^b_2)$ with cosine similarity greater than zero into a dataset $D_G$.
See Figure~\ref{fig:minedanalogies} for example transformation analogies.

We now use the dataset $D_G$ to train $G$.
For each quadruplet $(c^a_1, c^a_2, c^b_1, c^b_2)$, we feed $(c^a_1, c^b_1, c^b_2)$ to the generator.
Let $\hat{c}^a_2 = G ([ c^a_1, c^b_1, c^b_2 ])$ be the output of the generator.
We then minimize $\lambda L_{mse} ( \hat{c^a_2}, c^a_2) + L_{cls} ( W, \hat{c^a_2}, a)$, where: 
\begin{packed_enum}
\item $L_{mse} ( \hat{c^a_2}, c^a_2)$ is the mean squared error between the generator's output and the true target of the analogy $c^a_2$.
\item $L_{cls} ( W, \hat{c^a_2}, a)$ is the classification loss, where $W$ is the fixed linear classifier on the base classes learnt during representation learning, and $L_{cls} (W, x, y)$ is the log loss of the classifier $W$ on the example $(x,y)$.
\end{packed_enum}

\subsection{Using generated examples for low-shot learning}
\label{sec:usegen}
Our generated examples are unlikely to be as good as real examples, but should provide a useful bias to the classifier when only a few real examples are present.
Therefore we want to rely on generated examples only when the number of real examples is low.

Concretely, we have a hyperparameter $k$ (set through cross-validation), which is the minimum number of examples per novel category that we want to have.
If the actual number of real examples for a novel category, $n$, is less than $k$, then we additionally generate $k-n$ hallucinated examples.
To generate a synthetic example for a novel category $l$, we sample the feature vector of a ``seed'' example $\feat(x)$ from one of the $n$ real examples for this category, and a pair of cluster centroids $c^a_1, c^a_2$ from a base category $a$ chosen uniformly at random.
We then pass this triplet through $G$, and add the hallucinated feature vector $G([\feat(x), c^a_1, c^a_2])$ to our training set with label $l$.
We then train the logistic regression classifier on this mix of real and generated data in the usual manner. 

%% file: approach.tex
\section{Better representations for low-shot learning}
\label{sec:approach}
We now turn to the question of improving representation learning so as to enable better low-shot learning.
As described above, the learner consists of a \emph{feature extractor} $\feat$ and a \emph{classifier} $W$.
The goal of representation learning is a \emph{good feature extractor}: one that enables learning of effective classifiers from few examples.
Intuitively, our goal is to reduce the difference between classifiers trained on large datasets and classifiers trained on small datasets so that those trained on small datasets generalize better.

We first describe a proposal that encodes this goal in a loss that can be minimized during representation learning.
Then, we draw connections to several alternatives.

\subsection{Squared gradient magnitude loss (SGM)}
We assume that the classifier $W$ is linear, e.g., the last layer of a ConvNet.
Let $D$ denote a large labeled dataset of base class images.
Typically, training the feature extractor $\feat$ and the classifier $W$ on $D$ involves minimizing a classification objective with respect to $\feat$ and $W$:
\begin{equation}
\min_{W, \feat} L_D(\feat, W)  = \min_{W, \feat} \frac{1}{|D|} \sum_{(x,y) \in D} L_{cls}(W, \feat(x), y)
\end{equation}
where $L_{cls}(W, x, y)$  is the multiclass logistic loss on an example $x$ with label $y$ for a linear classifier $W$:
\begin{align}
L_{cls} ( W, x,y) &= - \log p_y(W, x) \\
p_k(W, x) &= \frac{\exp(w_k^Tx)}{\sum_j \exp(w_j^Tx)}.
\label{eq:pk}
\end{align}

We modify this training procedure as follows.
We \emph{simulate} low-shot learning experiments on the base classes by considering several tiny training sets $S \subset D$, $|S| \ll |D|$.
We then want to reduce the difference between classifiers trained on the large dataset $D$ (using the feature extractor $\feat$) and classifiers trained on these small datasets $S$.

The classifier trained on $D$ is just $W$.
Training a classifier on $S$ involves solving a minimization problem:
\begin{align}
 \min_V L_S(\feat,V) = \min_V \frac{1}{|S|} \sum_{(x,y) \in S} L_{cls}(V, \feat(x), y)
\end{align}

\begin{figure}
\centering
\includegraphics[width=0.6\linewidth]{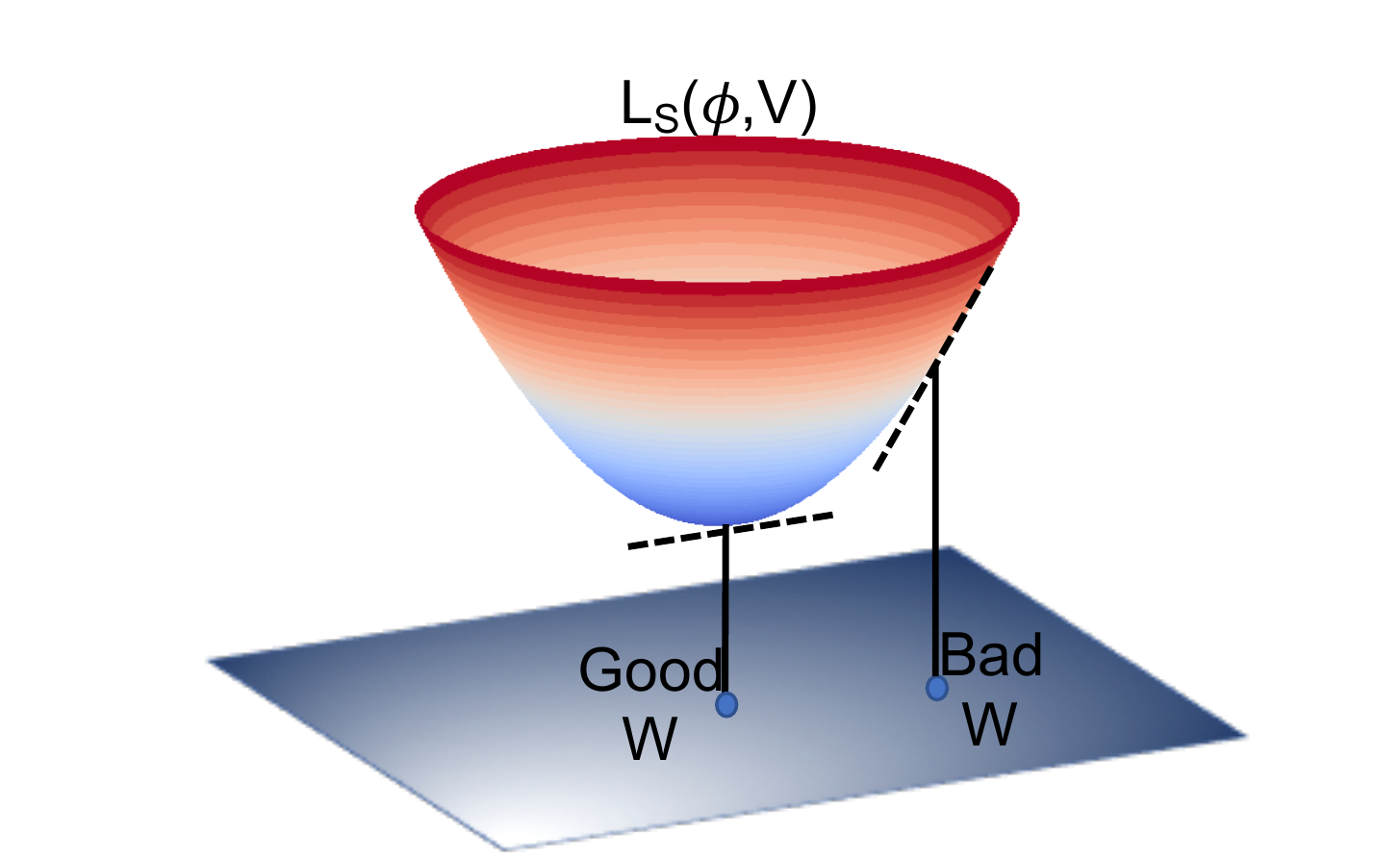}
\caption{Motivation for the SGM loss. We want to learn a representation $\feat$ such that the arg min of the small set training objective $L_S(\feat,V)$ matches $W$, the classifier trained on a large dataset $D$.}
\label{fig:sgmmotiv}
\end{figure}

We want the minimizer of this objective to match $W$.
In other words, we want $W$ to minimize $L_S(\feat, V)$.
$L_S(\feat, V)$ is convex in $V$ (Fig.~\ref{fig:sgmmotiv}), so a necessary and sufficient condition for this is that the gradient of $L_S(\feat, V)$ at $V=W$, denoted by $\nabla_V L_S(\feat, V) |_{\small{V=W}}$, is 0.
More generally, the closer $W$ is to the global minimum of $L_S(\feat, V)$, the lower the magnitude of this gradient.
Thus, we want to minimize:
\begin{equation}
\tilde{L}_S(\feat, W) = \|\nabla_V L_S(\feat, V) |_{V=W}\|^2
\end{equation}

The gradient $\nabla_V L_S(\feat, V)$ has a simple analytical form (see supplemental material for details\footnote{Supplemental material is available at \url{http://home.bharathh.info/lowshotsupp.pdf}} ):
\begin{align}
\nabla_V L_S(\feat, V) = [g_1(S,V), \ldots g_K(S,V)]\\
g_k(S, V) = \frac{1}{|S|}\sum_{(x,y) \in S} (p_k(V, \feat(x)) - \delta_{yk}) \feat(x) 
\end{align} 
where $K$ is the number of classes, $\delta_{yk}$ is 1 when $y = k$ and 0 otherwise, and $p_k$ is as defined in equation~\eqref{eq:pk}.

This leads to an analytical form for the function $\tilde{L}_S(\feat, W)$ : $\frac{1}{|S|^2}\sum_{k=1}^K \|\sum_{(x,y) \in S} (p_k(W, \feat(x)) - \delta_{yk}) \feat(x)\|^2$.
We use this analytical function of $W$ and $\feat$ as a loss.

We consider an extreme version of this loss where $S$ is a \emph{single example} $(x,y)$. In this case,
\begin{align}
\tilde{L}_S(\feat, W) &= \sum_{k=1}^K  (p_k(W, \feat(x)) - \delta_{yk})^2\| \feat(x) \|^2 \\
&= \alpha(W, \feat(x), y) \| \feat(x)\|^2.
 \label{eq:newloss2}
\end{align}
where $\alpha(W, \feat(x), y) = \sum_k (p_k(W, \feat(x)) - \delta_{yk})^2$ is a per-example weight that is higher for data points that are misclassified.
Thus the loss becomes a weighted $L_2$ regularization on the feature activations.

Our final loss, which we call SGM for Squared Gradient Magnitude, averages this over all examples in $D$.
\begin{equation}
L^{SGM}_D (\feat, W) = \frac{1}{|D|} \sum_{(x,y) \in D}  \alpha(W, \feat(x), y) \| \feat(x) \|^2
\end{equation} 

We train our feature representation by minimizing a straightforward linear combination of the SGM loss and the original classification objective.
\begin{equation}
\min_{W, \feat} L_D(\feat, W)+ \lambda L^{SGM}_D(\feat, W)
\label{eq:newloss}
\end{equation}
$\lambda$ is obtained through cross-validation.

%
%

\paragraph{Batch SGM.}
Above, we used singleton sets as our tiny training sets $S$.
An alternative is to consider every \emph{mini-batch} of examples $B$ that we see during SGD as $S$.
Hence, we penalize the squared gradient magnitude of the average loss over $B$, yielding the loss term: $\lambda \tilde{L}_B(\feat, W)$.
In each SGD iteration, our total loss is thus the sum of this loss term and the standard classification loss.
Note that because this loss is defined on mini-batches the number of examples per class in each mini-batch is a random variable.
Thus this loss, which we call ``batch SGM",  optimizes for an expected loss over a distribution of possible low-shot values $n$.

\subsection{Feature regularization-based alternatives}
In Eq.~\eqref{eq:newloss2}, it can be shown that $\alpha (W,\feat(x), y) \in [0,2]$ (see supplementary).
Thus, in practice, the SGM loss is dominated by $\| \feat(x) \|^2$, which is much larger.
This suggests a simple squared $L_2$ norm as a loss:
\begin{equation}
\min_{W, \feat} L_D (\feat, W) + \lambda \frac{1}{|D|} \sum_{(x,y) \in D}\|  \feat(x) \|^2.
\label{eq:l2}
\end{equation}
While $L_2$ regularization is a common technique, note that here we are regularizing the \emph{feature representation}, as opposed to regularizing the \emph{weight vector}.
Regularizing the feature vector norm has been a staple of unsupervised learning approaches to prevent degenerate solutions~\cite{RanzatoThesis2009}, but to the best of our knowledge it hasn't been considered in supervised classification. 

We can also consider other ways of regularizing the representation, such as an $L_1$ regularization:
\begin{equation}
\min_{W, \feat} L_D (\feat, W) + \lambda \frac{1}{|D|} \sum_{(x,y) \in D} \|  \feat(x) \|_1.
\label{eq:l2}
\end{equation}

We also evaluate other forms of feature regularization that have been proposed in the literature. 
The first  of these is dropout~\cite{HintonArxiv2012}, which was used in earlier ConvNet architectures~\cite{krizhevsky2012imagenet}, but has been eschewed by recent architectures such as ResNets~\cite{HeCVPR2016}.
Another form of feature regularization involves minimizing the correlation between the features~\cite{CheungICLR2015, CogswellICLR2016}.
We also compare to the multiverse loss~\cite{LittwinArxiv2015} which was shown to improve transfer learning performance.

\paragraph{Why should feature regularization help?}
When learning the classifier and feature extractor jointly, the feature extractor can choose to encode less discriminative information in the feature vector because the classifier can learn to ignore this information.
However, when learning new classifiers in the low-shot phase, the learner will not have enough data to identify discriminative features for the unseen classes from its representation. 
Minimizing the norm of the feature activations might limit what the learner can encode into the features, and thus force it to only encode useful information.

\subsection{Metric-learning based approaches}
A common approach to one-shot learning is to learn a good distance metric that generalizes to unseen classes.
We train a ConvNet with the triplet loss as a representative baseline method.
The triplet loss takes as input a triplet of examples $(x, x_+ , x_-)$, where $x$ and $x_+$ belong to the same category while $x_-$ does not:
\begin{align}
  & L_{\mathit{triplet}}(\feat(x), \feat(x_+), \feat(x_-))  = \\
  & \quad \max(  \|\feat(x_+) - \feat(x) \| - \| \feat(x_-) - \feat(x) \| + \gamma, 0). \nonumber
\end{align}
The loss encourages $x_-$ to be at least $\gamma$ farther away from $x$ than $x_+$ is.

%% file: experiments.tex
\section{Experiments and discussion}
\label{sec:experiments}

\subsection{Low-shot learning setup}
We use the ImageNet1k challenge dataset for experiments because it has a wide array of classes with significant intra-class variation.
We divided the 1000 ImageNet categories randomly into 389 base categories and 611 novel categories (listed in the supplementary material).

Many of the methods we evaluate have hyperparameters that need to be cross-validated.
Since we are interested in generalization to novel classes, we did not want to cross-validate on the same set of classes that we test on.
We therefore constructed two disjoint sets of classes by dividing the base categories into two subsets $C^1_{\base}$ (193 classes) and $C^2_{\base}$ (196 classes) and the novel categories into $C^1_{\novel}$ (300 classes) and $C^2_{\novel}$ (311 classes).
Then, for cross-validating hyperparameters, we provided the learner with $C^{\textit{cv}} = C^1_{\base} \cup C^1_{\novel}$ in the low-shot learning and testing phase, and evaluated its top-5 accuracy on the combined label set $C^{\textit{cv}}$.
The hyperparameter setting that gave the highest top-5 accuracy was then frozen.
We then conducted our final experiments using these hyperparameter settings by providing the learner with $C^{\textit{fin}} = C^2_{\base} \cup C^2_{\novel}$.
All reported numbers in this paper are on $C^{\textit{fin}}$.

Our test images are a subset of the ImageNet1k validation set: we simply restricted it to only include examples from the classes of interest ($C^{\textit{cv}}$ or $C^{\textit{fin}}$).
Performance is measured by top-1 and top-5 accuracy on the test images for each value of $n$ (number of novel examples per category).
We report the mean and standard deviation from 5 runs each using a different random sample of novel examples during the low-shot training phase.

To break down the final performance metrics, we report separately the average accuracy on the test samples from the novel classes and on all test samples.
While our focus is on the novel classes, we nevertheless need to ensure that good performance on novel classes doesn't come at the cost of lower accuracy on the base classes.

\subsection{Network architecture and training details}
For most of our experiments, we use a small ten-layer ResNet architecture~\cite{HeCVPR2016} as our feature extractor $\feat$ (details in supplementary material).
When trained on all 1000 categories of ImageNet, it gives a validation top-5 error rate of 16.7\% (center crop), making it similar to AlexNet~\cite{krizhevsky2012imagenet}.
We use this architecture because it's relatively fast to train (2 days on 4 GPUs) and resembles state-of-the-art architectures.
Note that ResNet architectures, as described in~\cite{HeCVPR2016}, do not use dropout.
Later, we show some experiments using the larger and deeper ResNet-50 architecture.

For all experiments on representation learning, except the triplet embedding, the networks are trained from scratch for 90 epochs on the base classes.
The learning rate starts at 0.1 and is divided by 10 every 30 epochs.
The weight decay is fixed at 0.0001.
For the triplet embedding, we first pretrain the network using a softmax classifier and log loss for 90 epochs, and then train the network further using the triplet loss and starting with a learning rate of 0.001.
We stop training when the loss stops decreasing (55 epochs).
This schedule is used because, as described in~\cite{RippelICLR2016}, triplet networks train slowly from scratch.

For methods that introduce a new loss, there is a hyperparameter that controls how much we weigh the new loss.
Dropout also has a similar hyperparameter that governs what fraction of activations are dropped.
We set these hyperparameters by cross-validation.

For our generator $G$, we use a three layer MLP with ReLU as the activation function.
We also add a ReLU at the end, since $\feat$ is known to be non-negative.
All hidden layers have a dimensionality of 512.

In the low-shot learning phase, we train the linear classifier using SGD for 10000 iterations with a mini-batch size of 1000. 
We cross-validate for the learning rate.

\subsection{Training with class imbalance}
The low-shot benchmark creates a heavily imbalanced classification problem.
During low-shot learning the base classes may have thousands of examples, while each novel class has only a few examples.
We use two simple strategies to mitigate this issue.
One, we oversample the novel classes when training the classifier by sampling uniformly over classes and then uniformly within each chosen class.
Two, we $L_2$ regularize the multi-class logistic classifier's weights by adding weight decay during low-shot learning.
We find that the weight of the classifier's $L_2$ regularization term has a large impact and needs to be cross-validated.

\subsection{Results}
\begin{figure*}[h!]
\centering
\includegraphics[width=0.30\textwidth]{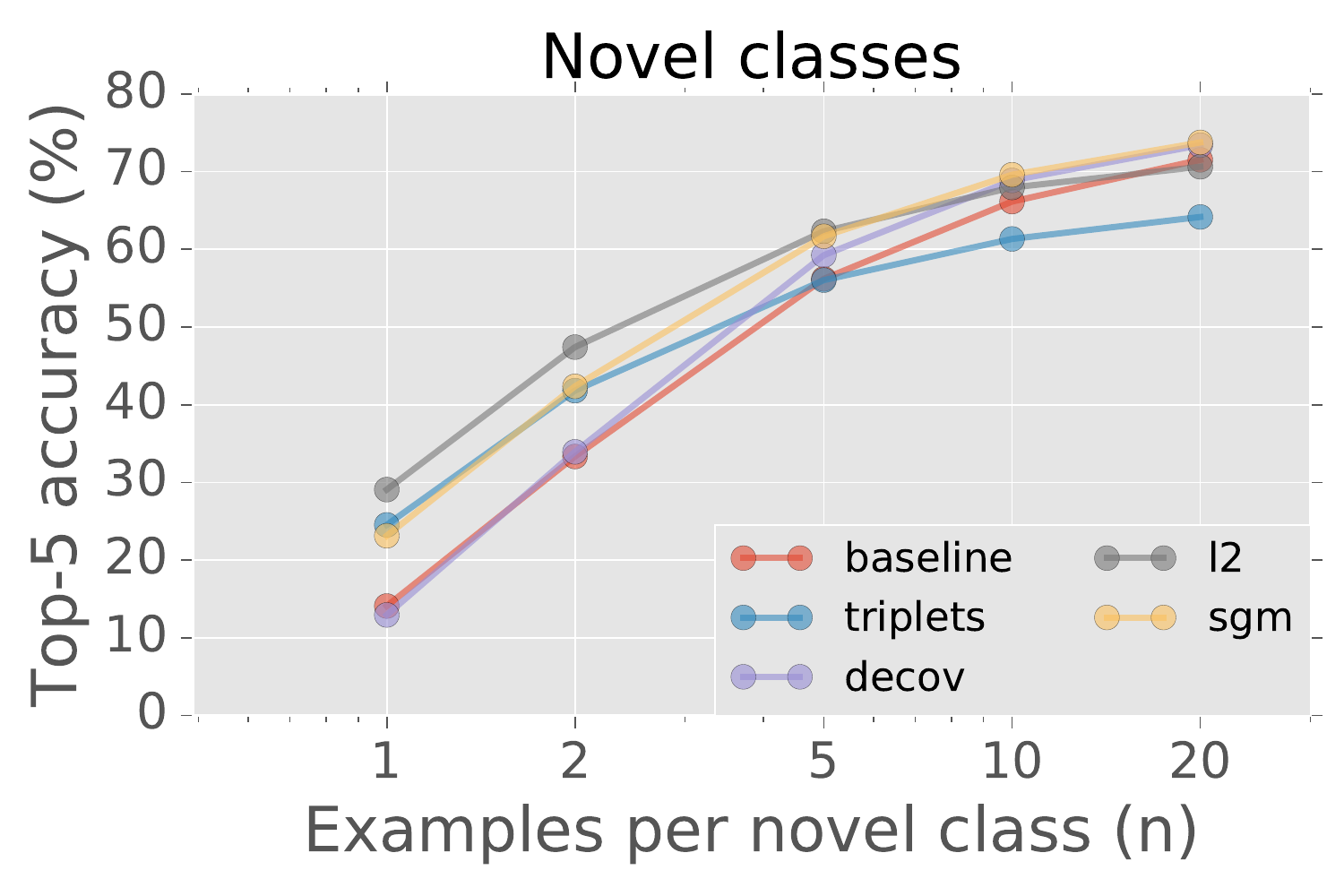}
\includegraphics[width=0.30\textwidth]{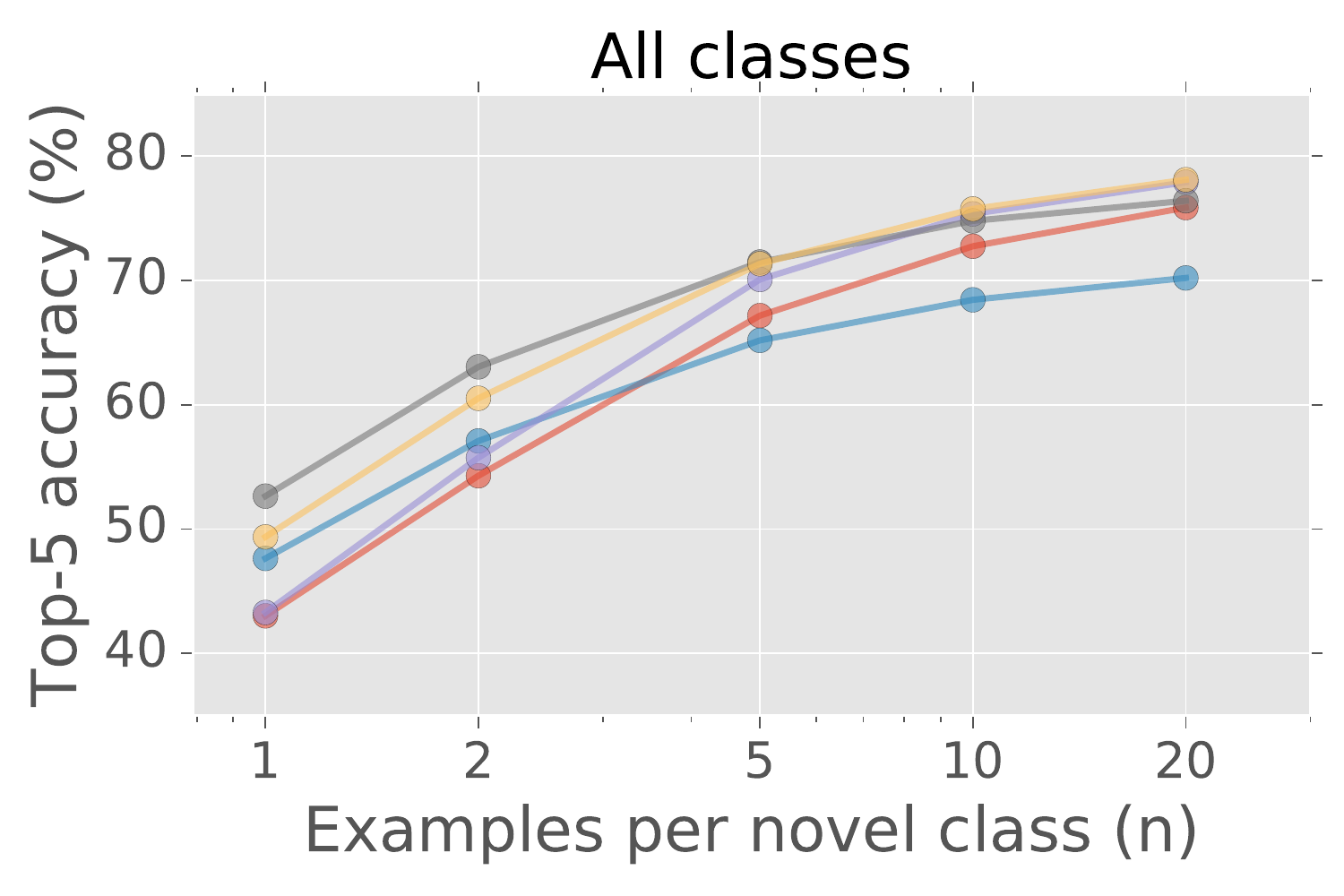}
\includegraphics[width=0.30\textwidth]{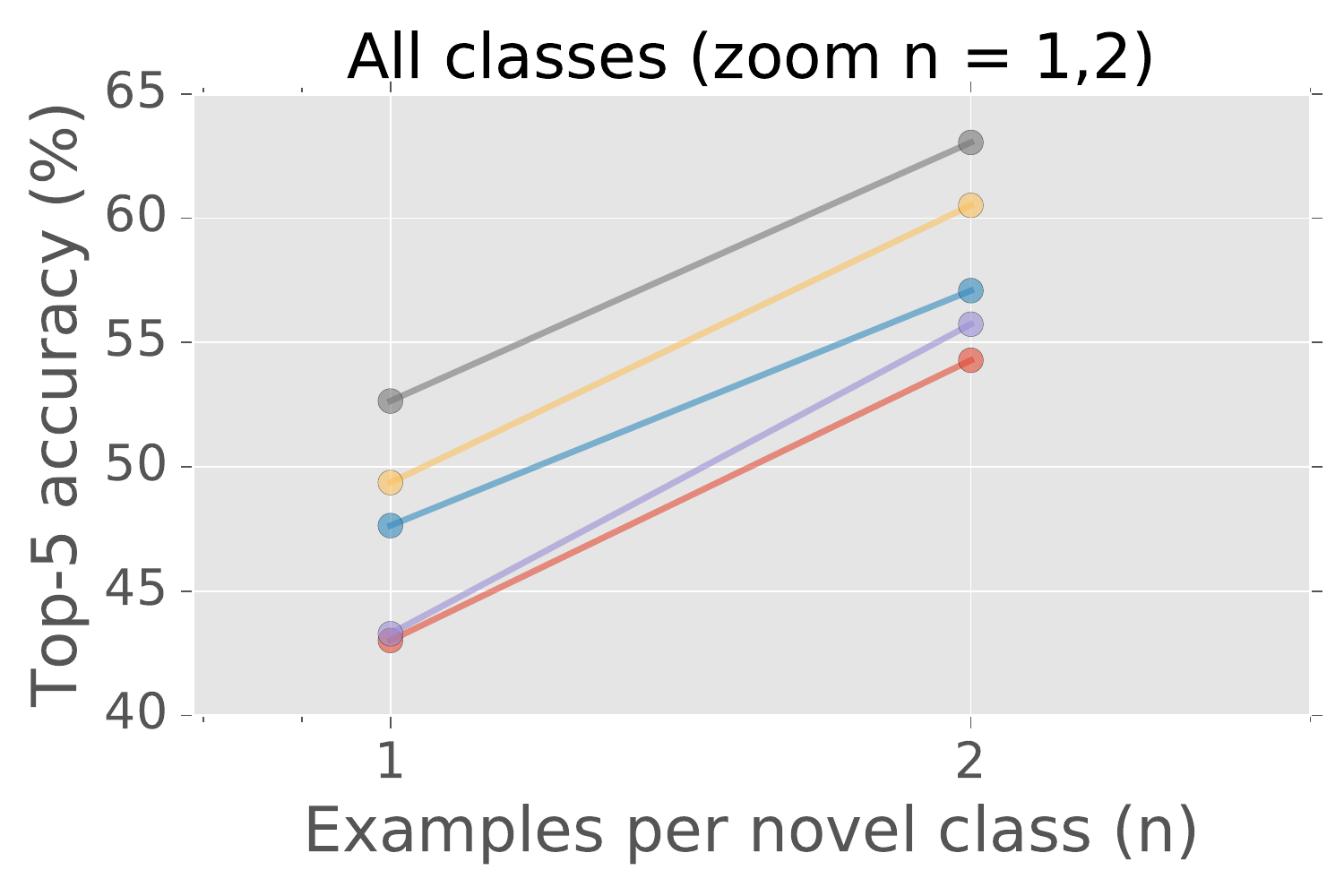}
\vspace{-1em}
\caption{\textbf{Representation learning comparison.} Top-5 accuracy on ImageNet1k val.
Top-performing feature regularization methods reduce the training samples needed to match the baseline accuracy by 2x. Note the different Y-axis scales.}
\label{fig:top5resultsrepr}
\end{figure*}

\textbf{Impact of representation learning.} We plot a subset of the methods\footnote{The subset reduces clutter, making the plots more readable. We omit results for Batch SGM, Dropout and $L_1$ because Batch SGM performs similarly to SGM and $L_2$, while $L_1$ and Dropout perform worse.} in Figure~\ref{fig:top5resultsrepr}, and show the full set of numbers in Tables~\ref{tab:novel} and~\ref{tab:all}.
The plots show the mean top-5 accuracy, averaged over 5 low-shot learning trials, for the novel classes, and over the combined set of novel and base classes.
The standard deviations are low (generally less than 0.5\%, see supplementary material) and are too small to display clearly as error bars.
Top-1 accuracy and numerical values are in the supplementary material.
We observe that:
\begin{packed_item}
\itemsep0.2em
\item When tested just on base classes, many methods perform similarly (not shown), but their performance differs drastically in the low-shot scenarios, especially for small $n$.
Thus, \emph{accuracy on base classes does not generalize to novel classes, especially when novel classes have very few training examples.}
\item Batch SGM, SGM, and $L_2$ are top performers overall with $L_2$ being better for small $n$.
They improve novel class accuracy by more than 10 points for small $n$ (1 or 2) and more than 3 points for $n>10$. 
$L_1$ also improves low-shot performance, but the gains are much smaller.
\item Dropout is on par with SGM for small $n$, but ends up being similar or worse than the baseline for $n\geq 5$ in terms of all class accuracy. 
Empirically, dropout also reduces feature norm, suggesting that implicit $L_2$ feature regularization might explain some of these gains.
\item Triplet loss improves accuracy for small $n$ but is 5 points worse than the baseline for $n=20$ in terms of all class accuracy.
While more sophisticated variants of the triplet loss may improve performance~\cite{RippelICLR2016}, feature regularization is both effective and much simpler. 
\item The decov loss~\cite{CogswellICLR2016} provides marginal gains for higher values of $n$ but is outperformed by the feature regularization alternatives. 
\end{packed_item}

As an additional experiment, we also attempted to finetune the baseline representation on all the base class examples and the small set of novel class examples. 
We found that this did not improve performance over the frozen representation (see Baseline-ft in Tables~\ref{tab:novel} and \ref{tab:all}).
This indicates that finetuning the representation is not only expensive, but also does not help in the low-shot learning scenario.

\begin{table}[t!]
\centering
\renewcommand{\arraystretch}{1.2}
\renewcommand{\tabcolsep}{1.2mm}
\resizebox{0.85\linewidth}{!}{
\input{plots/top5_novel.tex}}
\caption{Top-5 accuracy on only novel classes. Best are bolded and blue; the second best are italicized and red. $^*$Our methods.}
\label{tab:novel}
\end{table}
\begin{table}
\centering
\renewcommand{\arraystretch}{1.2}
\renewcommand{\tabcolsep}{1.2mm}
\resizebox{0.85\linewidth}{!}{
\input{plots/top5_all.tex}}
\caption{Top-5 accuracy on base and novel classes. Best are bolded and blue; the second best are italicized and red. $^*$Our methods.}
\label{tab:all}
\end{table}

\begin{figure*}[h!]
\centering
\includegraphics[width=0.3\textwidth]{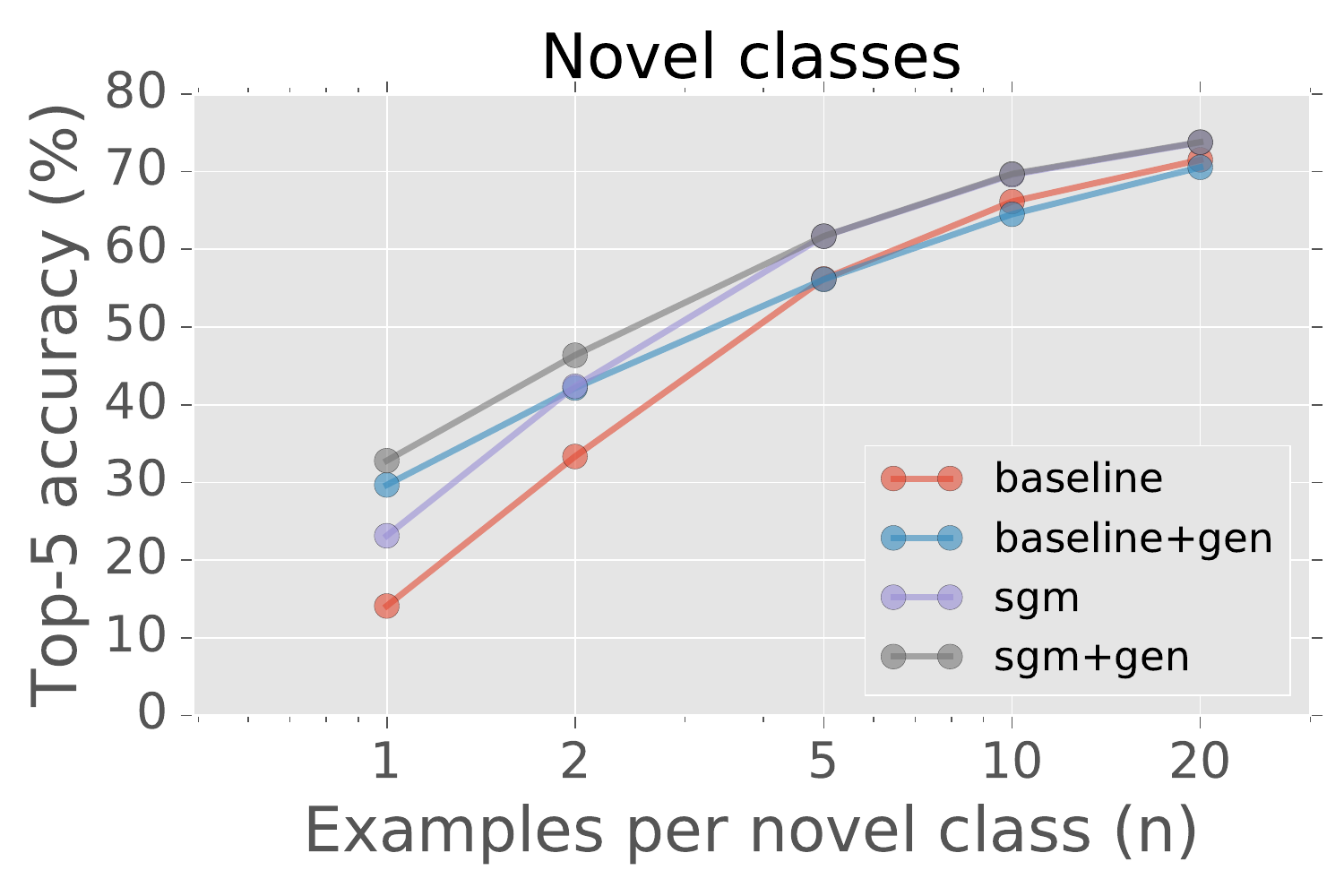}
\includegraphics[width=0.3\textwidth]{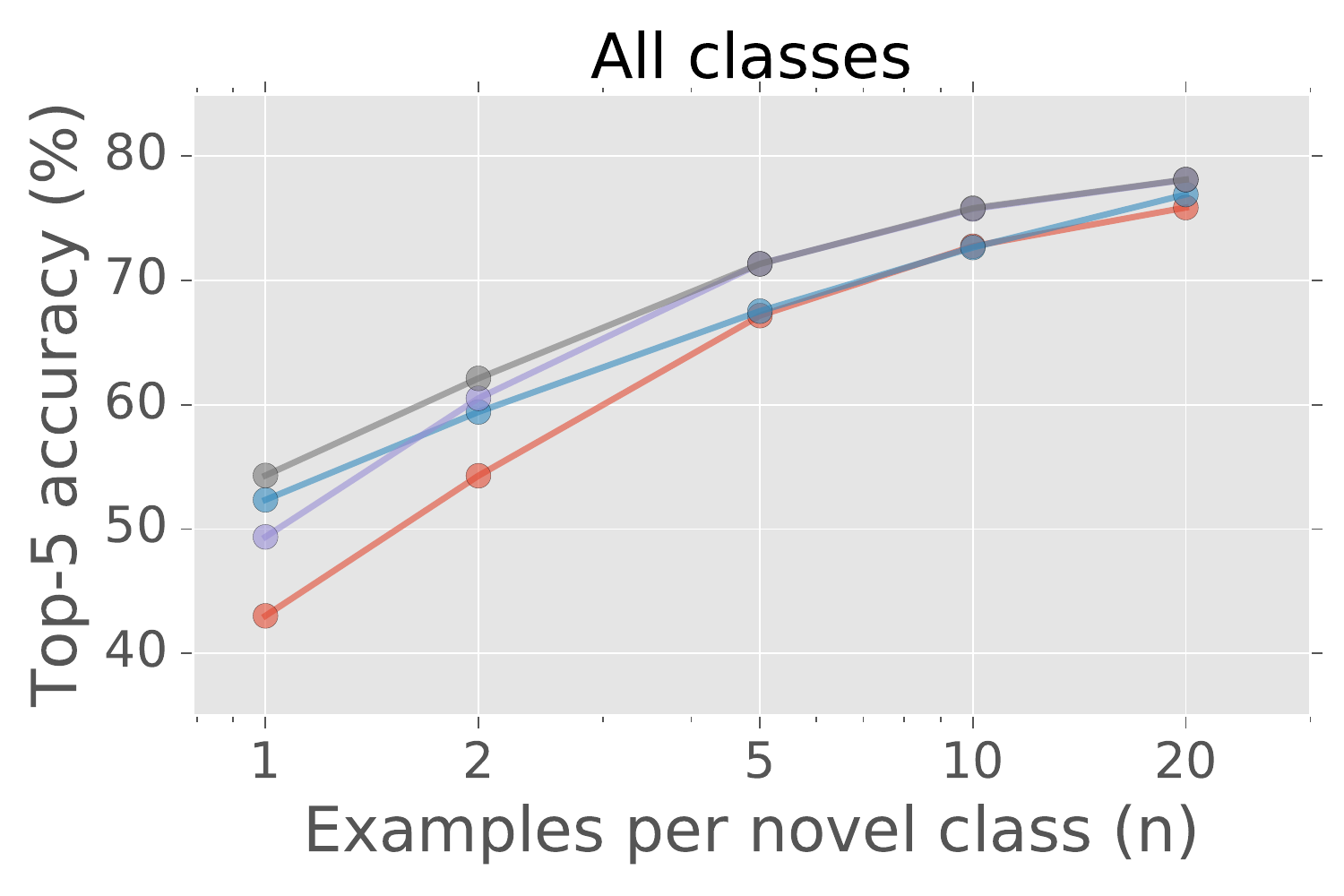}
\includegraphics[width=0.3\textwidth]{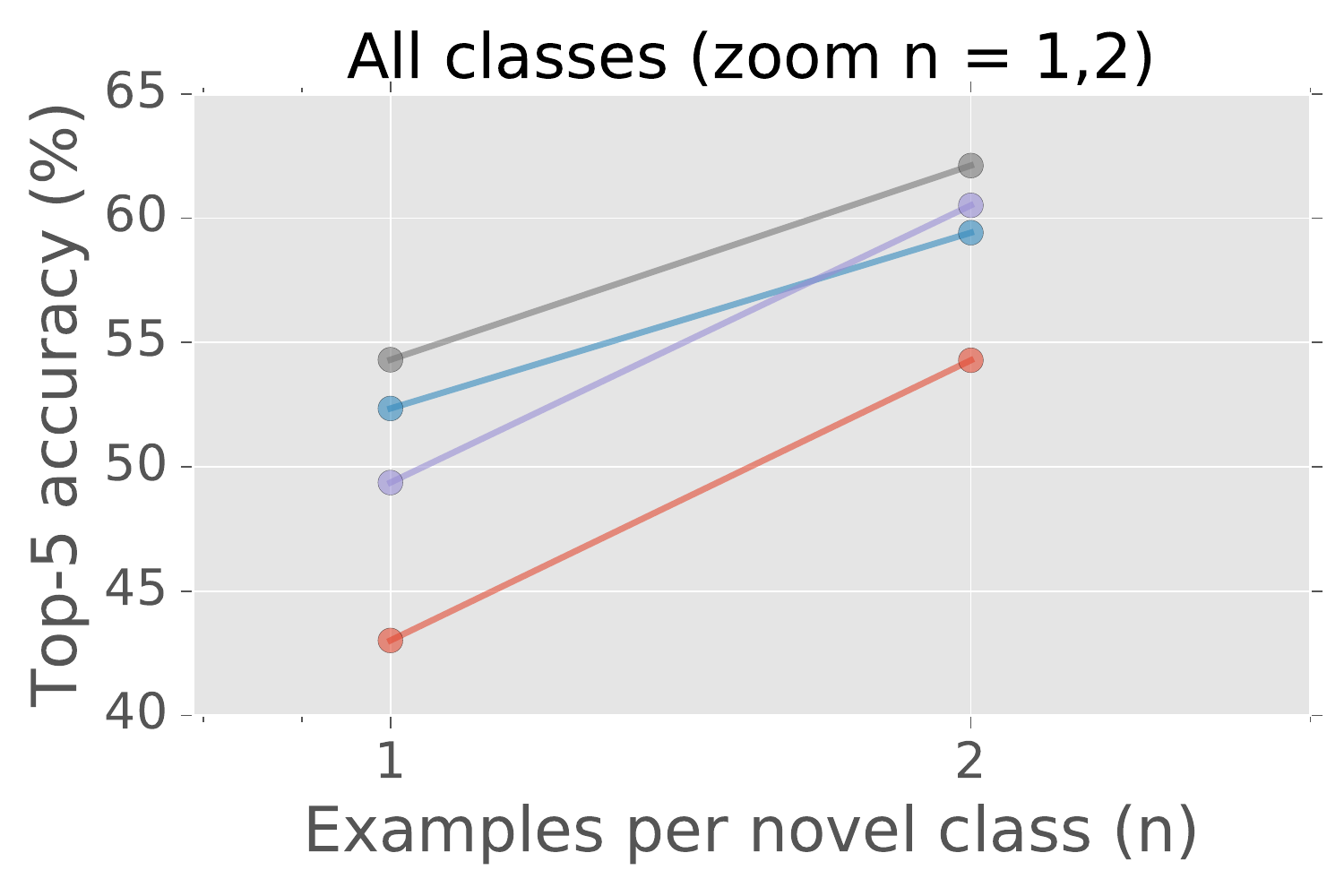}
\vspace{-1em}
\caption{\textbf{Comparisons with and without example generation.} Top-5 accuracy on ImageNet1k val. Note the different Y-axis scales.}
\label{fig:top5resultsgen}
\end{figure*}

\textbf{Impact of generation.} Figure~\ref{fig:top5resultsgen} shows the top-5 accuracies on novel classes and on base+novel classes for our generation method applied on top of the baseline representation and the SGM feature representation.
The numbers are in Tables~\ref{tab:novel} and~\ref{tab:all}.
Note that we only generate examples when $n<k$, with $k=20$ for baseline representations and 5 for SGM (see Section~\ref{sec:usegen}).
We observe that the generated examples provide a large gain of over 9 points for $n=1, 2$ on the novel classes for the baseline representation.
When using the SGM representation, the gains are smaller, but significant.

We also compared our generation strategy to common forms of data augmentation (aspect ratio and scale jitter, horizontal flips, and brightness, contrast and saturation changes). 
Data augmentation only provides small improvements (about 1 percentage point).
This confirms that our generation strategy produces more diverse and useful training examples than simple data augmentation.

\begin{figure*}[h!]
\centering
\includegraphics[width=0.3\textwidth]{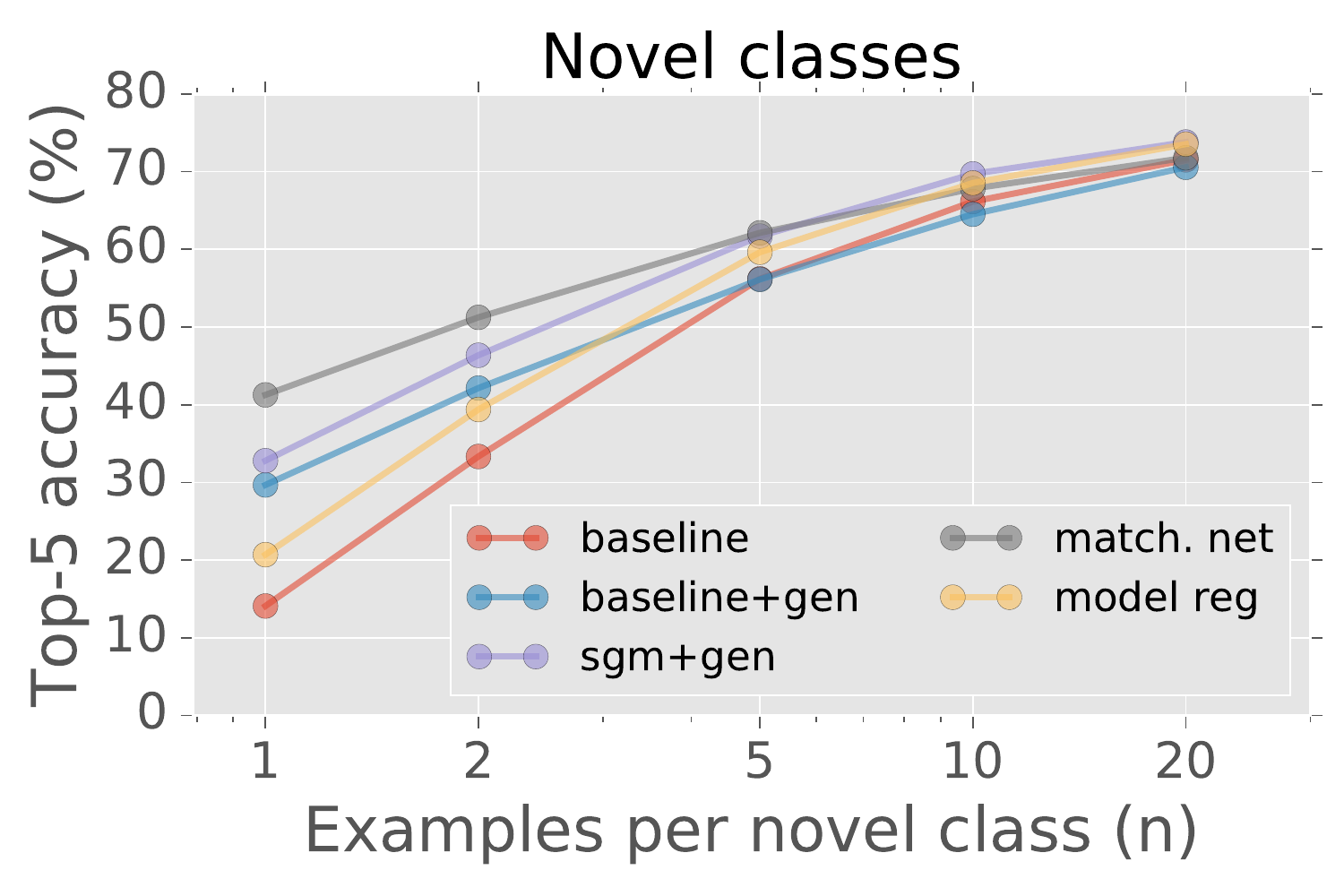}
\includegraphics[width=0.3\textwidth]{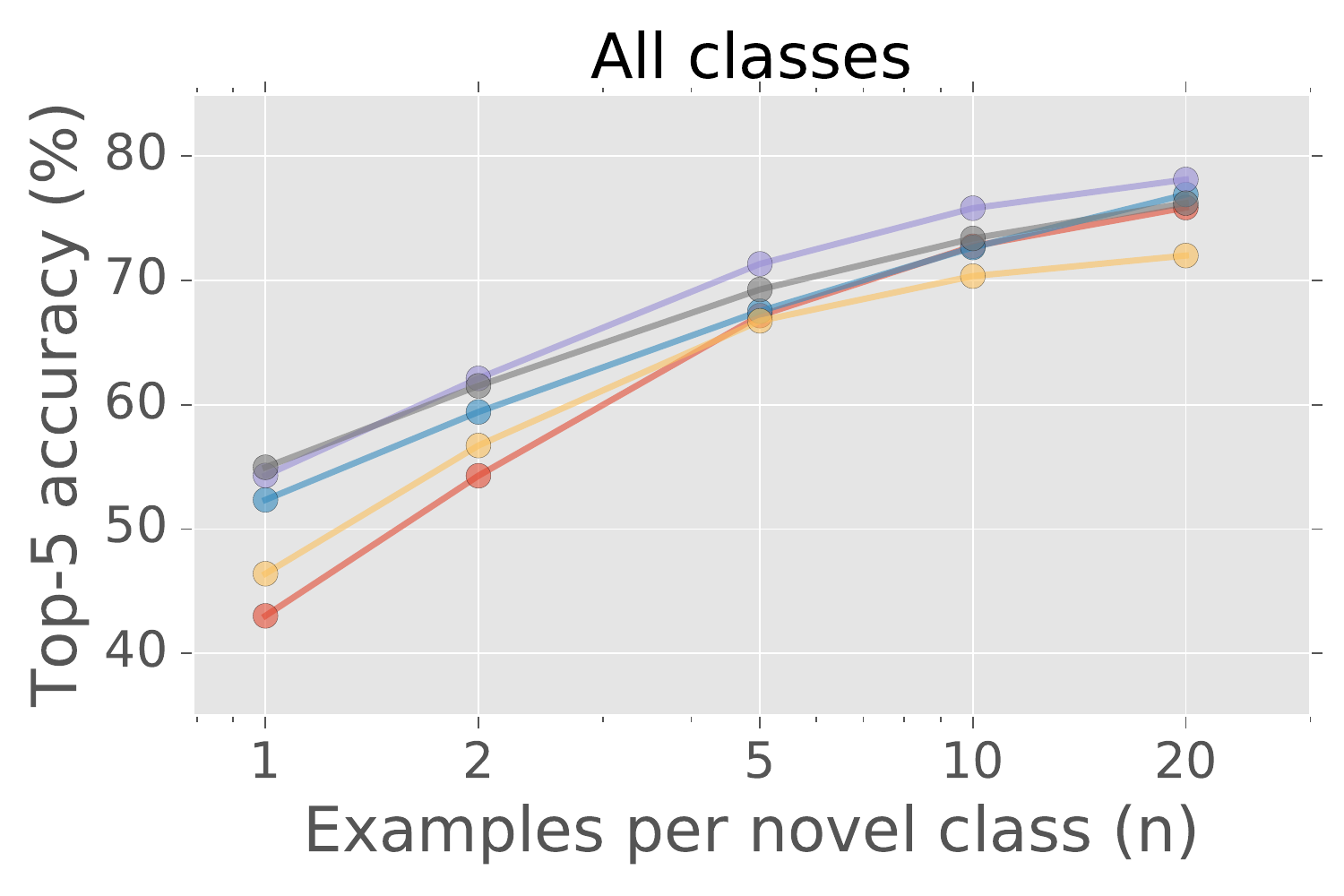}
\includegraphics[width=0.3\textwidth]{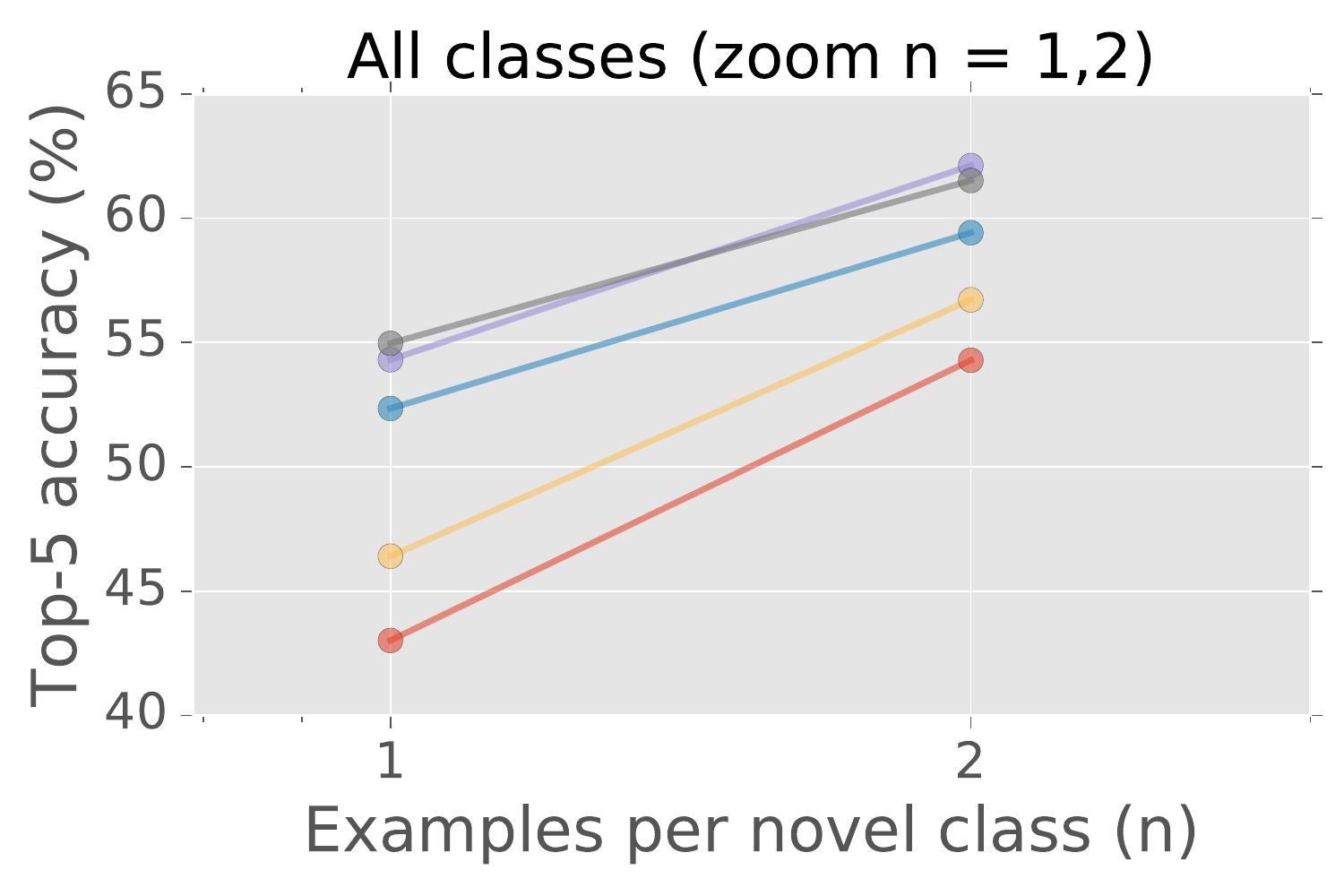}
\vspace{-1em}
\caption{\textbf{Comparison to recently proposed methods.} Top-5 accuracy on ImageNet1k val. Note the different Y-axis scales.}
\label{fig:top5resultsothers}
\end{figure*}

\textbf{Comparison to other low-shot methods.} We also compared to two recently proposed low-shot learning methods: matching networks~\cite{VinyalsNIPS2016} and model regression~\cite{WangECCV2016}. 
\textbf{Model regression} trains a small MLP to regress from the classifier trained on a small dataset to the classifier trained on the full dataset.
It then uses the output from this regressor to regularize the classifier learnt in the low-shot learning phase.
\textbf{Matching networks} proposes a nearest-neighbor approach that trains embeddings end-to-end for the task of low-shot learning.
We apply both these techniques on our baseline representation.

For both these methods, the respective papers evaluated on the novel classes only.
In contrast, real-world recognition systems will need to discriminate between data-starved novel concepts, and base classes with lots of data.
We adapt these methods to work with both base and novel classes as follows.
For model regression, we only use the model regressor-based regularization on the novel classes, with the other classifiers regularized using standard weight decay.
We use one-vs-all classifiers to match the original work.
 
Matching networks require the training dataset to be kept in memory during test time.
To make this tractable, we use 100 examples per class, with the novel classes correspondingly oversampled.
 
 Comparisons between these methods and our approach are shown in Figure~\ref{fig:top5resultsothers}.
 We find that model regression improves significantly over the baseline, but our generation strategy works better for low $n$.
 Model regression also hurts overall accuracy for high $n$.
 
 Matching networks work very well on novel classes.
 In terms of overall performance, they perform better than our generation approach on top of the baseline representation, but worse than generation combined with the SGM representation, especially for $n>2$.
 Further, matching networks are based on nearest neighbors and keep the entire training set in memory, making them much more expensive at test time than our simple linear classifiers.

\textbf{Deeper networks.}
 We also evaluated our approach on the ResNet-50 network architecture to test if our conclusions extend to deeper convnets that are now in use (Tables~\ref{tab:novel} and~\ref{tab:all}).
 First, even with the baseline representation and without any generation we find that the deeper architecture also leads to improved performance in all low-shot scenarios.
 However, our SGM loss and our generation strategy further improve this performance.
 Our final top-5 accuracy on novel classes is still more than 8 points higher for $n=1, 2$, and our overall accuracy is about 3 points higher, indicating that our contributions generalize to deeper and better models.

%% file: plots/top5_novel.tex
\begin{tabular}{llccccc}
\toprule
Representation & Lowshot phase & n=1 & 2 & 5 & 10 & 20 \\
\midrule
\emph{ResNet-10} & & & & & & \\
\;Baseline & Classifier & 14.1 & 33.3 & 56.2 & 66.2 & 71.5 \\
\;Baseline & Generation$^*$ + Classifier & 29.7 & 42.2 & 56.1 & 64.5 & 70.0 \\
\;SGM$^*$ & Classifier & 23.1 & 42.4 & \secondrank{61.7} & 69.6 & \secondrank{73.8} \\
\;SGM$^*$ & Generation$^*$ + Classifier & \secondrank{32.8} & 46.4 & \secondrank{61.7} & \secondrank{69.7} & \secondrank{73.8} \\
\;Batch SGM$^*$ & Classifier & 23.0 & 42.4 & 61.9 & \firstrank{69.9} & \firstrank{74.5} \\
\;L1$^*$ & Classifier & 20.8 & 40.8 & 59.8 & 67.5 & 71.6 \\
\;L2$^*$ & Classifier & 29.1 & \secondrank{47.4} & \firstrank{62.3} & 68.0 & 70.6 \\
\;Triplets & Classifier & 24.5 & 41.8 & 56.0 & 61.3 & 64.2 \\
\;Dropout~\cite{HintonArxiv2012} & Classifier & 26.8 & 43.9 & 59.6 & 66.2 & 69.5 \\
\;Decov~\cite{CogswellICLR2016} & Classifier & 13.0 & 33.9 & 59.3 & 68.9 & 73.4 \\
\;Multiverse~\cite{LittwinArxiv2015}& Classifier & 13.7 & 30.6 & 52.5 & 63.8 & 71.1 \\
\;Baseline & Data augmentation & 16.0 & 31.4 & 52.7 & 64.4 & 71.8 \\
\;Baseline & Model Regression~\cite{WangECCV2016} & 20.7 & 39.4 & 59.6 & 68.5 & 73.5 \\
\;Baseline & Matching Network~\cite{VinyalsNIPS2016} & \firstrank{41.3} & \firstrank{51.3} & \secondrank{62.1} & 67.8 & 71.8 \\
\;Baseline-ft & Classifier & 12.5 & 29.5 & 53.1 & 64.6 & 70.4 \\
\midrule
\emph{ResNet-50} & & & & & & \\
\;Baseline  & Classifier & 28.2 & 51.0 & 71.0 & \secondrank{78.4} & \secondrank{82.3} \\
\;Baseline  & Generation$^*$ + Classifier & \secondrank{44.8} & \firstrank{59.0} & \secondrank{71.4} & 77.7 & \secondrank{82.3} \\
\;SGM$^*$  & Classifier & 37.8 & 57.1 & \firstrank{72.8} & \firstrank{79.1} & \firstrank{82.6} \\
\;SGM$^*$  & Generation$^*$ + Classifier & \firstrank{45.1} & \secondrank{58.8} & \secondrank{72.7} & \firstrank{79.1} & \firstrank{82.6} \\
\bottomrule
\end{tabular}

%% file: plots/top5_all.tex
\begin{tabular}{llccccc}
\toprule
Representation & Lowshot phase & n=1 & 2 & 5 & 10 & 20 \\
\midrule
\emph{ResNet-10} & & & & & & \\
\; Baseline & Classifier & 43.0 & 54.3 & 67.2 & 72.8 & 75.9 \\
\; Baseline & Generation$^*$ + Classifier & 52.4 & 59.4 & 67.5 & 72.6 & 76.9 \\
\; SGM$^*$ & Classifier & 49.4 & 60.5 & \secondrank{71.3} & \firstrank{75.8} & \secondrank{78.1} \\
\; SGM$^*$ & Generation$^*$ + Classifier & \secondrank{54.3} & \secondrank{62.1} & \secondrank{71.3} & \firstrank{75.8} & \secondrank{78.1} \\
\; Batch SGM$^*$ & Classifier & 49.3 & 60.5 & 71.4 & \secondrank{75.8} & \firstrank{78.5} \\
\; L1$^*$ & Classifier & 47.1 & 58.5 & 69.2 & 73.7 & 76.1 \\
\; L2$^*$ & Classifier & 52.7 & \firstrank{63.0} & \firstrank{71.5} & 74.8 & 76.4 \\
\; Triplets & Classifier & 47.6 & 57.1 & 65.2 & 68.4 & 70.2 \\
\; Dropout~\cite{HintonArxiv2012} & Classifier & 50.1 & 59.7 & 68.8 & 72.7 & 74.7 \\
\; Decov~\cite{CogswellICLR2016} & Classifier & 43.3 & 55.7 & 70.1 & 75.4 & 77.9 \\
\; Multiverse~\cite{LittwinArxiv2015} & Classifier & 44.1 & 54.2 & 67.0 & 73.2 & 76.9 \\
\; Baseline & Data Augmentation & 44.9 & 54.0 & 66.4 & 73.0 & 77.2\\
\; Baseline & Model Regression~\cite{WangECCV2016} & 46.4 & 56.7 & 66.8 & 70.4 & 72.0 \\
\; Baseline & Matching Network~\cite{VinyalsNIPS2016} & \firstrank{55.0} & 61.5 & 69.3 & 73.4 & 76.2 \\
\; Baseline-ft & Classifier & 41.7 & 51.7 & 65.0 & 71.2 & 74.5 \\
\midrule
\emph{ResNet-50} & & & & & & \\
\; Baseline  & Classifier & 54.1 & 67.7 & \secondrank{79.1} & \secondrank{83.2} & \firstrank{85.4} \\
\; Baseline  & Generation$^*$ + Classifier & \secondrank{63.1} & \firstrank{71.5} & 78.8 & 82.6 & \firstrank{85.4} \\
\; SGM$^*$  & Classifier & 60.0 & \secondrank{71.3} & \firstrank{80.0} & \firstrank{83.3} & \secondrank{85.2} \\
\; SGM$^*$ & Generation$^*$ + Classifier & \firstrank{63.6} & \firstrank{71.5} & \firstrank{80.0} & \firstrank{83.3} & \secondrank{85.2} \\
\bottomrule
\end{tabular}

%% file: conclusion.tex
\section{Conclusion}
This paper proposes (1) a low-shot recognition benchmark of realistic complexity, (2) the squared gradient magnitude (SGM) loss that encodes the end-goal of low-shot learning, and (3) a novel way of transferring modes of variation from base classes to data-starved ones.
Source code and models are available at: \url{https://github.com/facebookresearch/low-shot-shrink-hallucinate}.